# Irrelevance and Independence Relations in Quasi-Bayesian Networks

Fabio Cozman*
Escola Politécnica, University of São Paulo
fgcozman@usp.br — http://www.cs.cmu.edu/~fgcozman/home.html

## Abstract

*This paper analyzes irrelevance and independence relations in graphical models associated with convex sets of probability distributions (called Quasi-Bayesian networks). The basic question in Quasi-Bayesian networks is, How can irrelevance/independence relations in Quasi-Bayesian networks be detected, enforced and exploited? This paper addresses these questions through Walley's definitions of irrelevance and independence. Novel algorithms and results are presented for inferences with the so-called natural extensions using fractional linear programming, and the properties of the so-called type-1 extensions are clarified through a new generalization of d-separation.*

## 1 INTRODUCTION

The theory of convex sets of distributions, variously called the theory of imprecise probabilities [28] or Quasi-Bayesian theory [14], is appropriate for robustness analysis [1, 19, 29] and for representation of imprecise/incomplete beliefs and opinions [20].

*Quasi-Bayesian networks* are multivariate structures that represent convex sets of joint distributions by directed acyclic graphs [3, 9, 27]. The key technical problem in Quasi-Bayesian networks is how to detect, enforce and exploit irrelevance and independence relations. The goal of this paper is to present novel results and algorithms that address these questions. This paper adopts Walley's definitions of irrelevance and investigates two different methods to generate infereces from Quasi-Bayesian network: inferences from *type-1 extensions* (Section 4), and inferences from *natural extensions* (Section 5).

*This research was conducted while the author was with the Robotics Institute, Carnegie Mellon University. The project was partially supported by NASA under Grant NAGW-1175; the author was supported under a scholarship from CNPq, Brazil.

The overall contribution of this paper is a theory of locally defined Quasi-Bayesian networks that display the same flexibility and representational power of standard Bayesian networks. The results in this paper state the conditions that must be required or enforced to express judgements of irrelevance/independence through Quasi-Bayesian networks.

## 2 BACKGROUND MATERIAL

### 2.1 GRAPHICAL MODELS

A popular graphical representation of probabilistic models in AI is the *Bayesian network* formalism, where a directed acyclic graph is used to specify a joint distribution over a set of variables $\tilde{X}$ [18]. Each node of a Bayesian network is associated with a variable $X_i$; the parents of $X_i$ are denoted by $\mathrm{pa}(X_i)$. This paper deals with variables with a finite set of values.

Each variable in a Bayesian network is associated with a conditional distribution $p(X_i|\mathrm{pa}(X_i))$. Such a graphical structure defines a unique joint probability distribution through the following expression [22]:

$$p(\tilde{X}) = \prod_i p(X_i|\mathrm{pa}(X_i)). \qquad (1)$$

Inferences with Bayesian networks usually involve the calculation of the posterior marginal for a *queried* variable $X_q$ given evidence $E$ [18].

Bayesian networks represent many independence relations among the variables in the network. These relations can be analyzed through the concept of *d-separation*: if $\tilde{Y}$ d-separates $X$ from $\tilde{Z}$, then $X$ and $\tilde{Z}$ are independent given events defined by $\tilde{Y}$ [22, page 117].

One difficulty with Bayesian networks is the requirement that all probability distributions must be precisely specified. Several non-probabilistic attempts have been made to relax the requirements of Bayesian networks through alternative theories of inference [12, 25, 26], or through interval-valued probabilities [2, 11, 16, 15]. Interval representations have two problems. First, it is not always possible to apply Bayes rule to an interval-valued distribution and obtain an interval-valued posterior distribution [7, 15]. Second, there is no unique, accepted way to define independence for interval-valued distributions [6].

Closed convex sets of distributions are also models for imprecision in probability values [1, 15, 24, 28]. Closed convex sets of distributions have several advantages when compared to interval-valued probability because conditionalization and independence can be defined without technical difficulties. In this paper, closed convex sets of distributions are employed as fundamental entities that reflect perturbations and imprecision about stochastic phenomena.

One axiomatization of closed convex sets of distributions that is particularly concise and powerful is the *Quasi-Bayesian* theory of Giron and Rios [14]. This theory is summarized in the next section; several recent definitions and results, not present in the original theory by Giron and Rios, are incorporated in the presentation.

## 2.2  QUASI-BAYESIAN THEORY

Quasi-Bayesian theory [14] uses convex sets of distributions to represent beliefs and to evaluate decisions.

**Credal sets**  The convex set of distributions maintained by an agent is called the agent's *credal* set, and its existence is postulated on the grounds of axioms about preferences [14].[1] To simplify terminology, the term *credal set* is used only when it refers to a set of distributions containing more than one element. A closed convex set of joint distributions is called a *joint credal set*.

This paper deals with credal sets that are defined as the convex hull of a finite number of probability distributions; such *finitely generated credal sets* are polytopes in the space of probability distributions.

**Lower and upper values**  Given a convex set $K$ of probability distributions, a probability interval can be created for every event $A$ by defining lower and upper bounds:
$$\underline{p}(A) = \inf_{p \in K} p(A), \qquad \overline{p}(A) = \sup_{p \in K} p(A).$$

---
[1]An introduction to technical aspects of Quasi-Bayesian theory, with a larger list of references, can be found at http://www.cs.cmu.edu/~fgcozman/qBayes.html.

Lower and upper expectations for a function $f(X)$ are defined as ($E_p[f]$ is the expectation of the function $f$):
$$\underline{E}[f] = \inf_{p \in K} E_p[f], \qquad \overline{E}[f] = \sup_{p \in K} E_p[f].$$
There is a one-to-one correspondence between lower (or upper) expectations and credal sets. Given a credal set, the set of all lower expectations for all arbitrary functions $f(X)$ is unique, and vice-versa.

**Conditionalization**  Convex sets of conditional distributions are used to represent conditional beliefs. Inference is performed by applying Bayes rule to each distribution in a jont credal set. The posterior credal set is the union of all posterior distributions obtained in this process, and the vertices of the posterior credal set are obtained by applying Bayes rule to all vertices of the joint credal set [28]. Denote by $K(X|Y)$ the collection of credal sets $K(X|A)$ for all events $A$ defined by a variable $Y$.

**Independence**  There is no unique way to define independence relations with credal sets; in the most in-depth study of this matter, Campos and Moral have reviewed five different possible types of independence [10]. The results presented in this paper adopt Walley's definition of independence [28]. Walley's original definition is stated in terms of lower expectations; to develop a theory of convex sets of distributions, it is important to recast Walley's definition using credal sets as follows.

Consider sets of variables $X$, $Y$ and $Z$ and the credal sets $K(X,Y,Z)$, $K(X|Z)$, $K(Y|Z)$, $K(X|Y,Z)$ and $K(Y|X,Z)$. Note that distributions in $K(X|Z)$ and $K(X|Y,Z)$ are defined over the same algebra of events once $Y$ and $Z$ are fixed; likewise, distributions in $K(Y|Z)$ and $K(Y|X,Z)$ are defined over the same algebra of events once $X$ and $Z$ are fixed.

**Definition 1** *Variables $\tilde{Y}$ are irrelevant to $\tilde{X}$ given $\tilde{Z}$ if $K(\tilde{X}|\tilde{Z})$ is equal to $K(\tilde{X}|\tilde{Y},\tilde{Z})$ regardless of the value of $\tilde{Y}, \tilde{Z}$. Variables $\tilde{X}$ and $\tilde{Y}$ are independent given $\tilde{Z}$ if $\tilde{X}$ is irrelevant to $\tilde{Y}$ given $\tilde{Z}$ and $\tilde{Y}$ is irrelevant to $\tilde{X}$ given $\tilde{Z}$. If $\tilde{Z}$ is empty, suppress the "given $\tilde{Z}$" from this definition.*

This concept of independence does not imply that joint credal sets contain only joint distributions with independent marginals, nor does it imply uniqueness for the joint credal set [28, Chapter 9].

## 3  LOCALLY DEFINED QUASI-BAYESIAN NETWORKS

This section defines Quasi-Bayesian networks that are generated from local models associated to a directed

Quasi-Bayesian Networks    91...-

acyclic graph:

**Definition 2** *A locally defined Quasi-Bayesian network is a directed acyclic graph associated with: (1) either a single conditional distribution $p(X_i|pa(X_i))$ or a local credal set $K(X_i|pa(X_i))$ for each variable $X_i$, (2) a collection of irrelevance relations, and (3) a method for the combination of local credal sets.*

A joint credal set that satisfies all constraints and relations in a Quasi-Bayesian network is called an *extension* of the network.

The rationale for this definition is as follows. In a standard Bayesian network, irrelevance and independence constraints are implicit in Expression (1); this expression guarantees that a variable is independent of all its non-descendants given its parents [22, page 119]. There is no analogue to Expression (1) in Quasi-Bayesian networks. Many extensions may satisfy all graphical d-separation relations in a network. It seems more appropriate to ask a decision maker to explicitly indicate which qualitative constraints are to be enforced in a Quasi-Bayesian network, and to ask for irrelevance constraints instead of independence constraints, because irrelevance and independence are not equivalent in Quasi-Bayesian models (Section 2.2).

The key fact is that a directed acyclic graph and a collection of local credal sets may admit more than one extension; the next sections investigate two important types of extension.

## 4  TYPE-1 EXTENSION

The most popular type of extension disussed in the literature is the type-1 extension [4, 27]. A type-1 extension is the convex hull of all the joint distributions formed by cross-multiplication of *extreme points* of local credal sets; consequently, a type-1 extension is the largest joint credal set where all extreme points satisfy Expression (1).

The appeal of type-1 extensions comes from their intuitive similarity with standard Bayesian networks. The following theorem formalizes this intuition using Walley's definition of independence:

**Theorem 1** *Every graphical d-separation relation in a Quasi-Bayesian network corresponds to a valid conditional independence relation in the type-1 extension of the network. (Proof in Appendix A.1.)*

This result demonstrates that type-1 extensions mimic the properties of standard Bayesian networks as independence-maps [22, page 119]. The theorem also complements results by Cano et al. [4]. They give conditions on independence concepts that satisfy d-separation in type-1 extensions, but they do not provide any definition of independence to illustrate their result. The theorem demonstrates that Walley's independence relations exhibit the desired correspondence with d-separation.

D-separation has important algorithmic consequences. Graphical operations that are guaranteed by d-separation can be performed in a type-1 extension. In particular, consider a query involving a variable $X_q$ and evidence $E$. All variables that do not affect computation of $p(X_q|E)$ can be detected through d-separation computations [13]. This greatly reduces the computational effort in Quasi-Bayesian inferences with type-1 extensions both for exact (enumeration) and approximate (sampling, iterative) algorithms [9]. The theorem in this section completes that investigation with a formal proof that d-separation can (and should) be used to handle type-1 extensions.

## 5  NATURAL EXTENSION

Type-1 extensions are *not* the only possible extension of a locally defined Quasi-Bayesian network. The *natural extension* of the network is the largest set of joint distributions compatible with local credal sets and irrelevance relations in the network. This terminology has been suggested by Walley [28, pages 453, 455], who explores properties of natural extensions but does not focus on multivariate structures.

A Quasi-Bayesian network is defined by quantitative constraints on probability values and by qualitative statements of irrelevance and independence. The quantitative constraints that define a credal set $K(X_i|\text{pa}(X_i))$ are denoted by $C_l[p(X_i|\text{pa}(X_i))]$.

The objective of this section is to investigate and exploit the representation of qualitative statements of irrelevance and independence in natural extensions, particulary statements that involve variables and their nondescendants. Many different natural extensions can be created for a given directed acyclic graph through different statements of irrelevance (Section 3).

The algorithms focus on irrelevance and independence conditional on the nondescendants of a node. This strategy follows common practice in Bayesian networks, which are based on the agreement between d-separation and irrelevance/independence [22]; for natural extension, this strategy has a simple justification as follows. When stating irrelevance/independence relations among variables, it is important to guarantee that a natural extension can actually be constructed. incompatible relations can lead to an empty natural extension. One strategy that always produces valid natural extensions is to rely on graphical d-separations as the source of irrelevance/independence relations, because there is always at least one standard Bayesian joint distribution that complies with all constraints. This rationale suggests that irrelevance/independence relations among variables and their nondescendants are of primary interest.

## 5.1 SPECIFYING CONDITIONAL CREDAL SETS SEPARATELY

The following algorithms assume that constraints on conditional distributions are defined separately for each value of the variable's parents. This means that, for any variable $X_i$, the constraints $C_l[p(X_i|[\text{pa}(X_i)]_{k_1})]$ do not interfere with the constraints for $C_l[p(X_i|[\text{pa}(X_i)]_{k_2})]$ when $k_1 \neq k_2$. This restriction makes sense both during elicitation of models and representation of constraints, and the following derivations exploit this restriction to generate inference algorithms.

Consider first the quantitative constraints $C_l[p(X_i|[\text{pa}(X_i)]_k)]$. Because all local credal sets have a finite number of vertices, all constraints $C_l[p(X_i|[\text{pa}(X_i)]_k)]$ are linear in $p(X_i|[\text{pa}(X_i)]_k)$. Because the value of $\text{pa}(X_i)$ is fixed in every constraint, all constraints are of the form:

$$\sum_{j=1}^{|\tilde{X}_i|} \gamma_{ijkl} p(X_i = X_{ij}, [\text{pa}(X_i)]_k) \leq \gamma_{i0kl} p([\text{pa}(X_i)]_k), \quad (2)$$

where $\gamma_{ijkl}$ are constants that define the local credal sets. Note that these constraints are linear in $p(\tilde{X})$, because $p(X_i, \text{pa}(X_i))$ and $p(\text{pa}(X_i))$ are summations over $p(\tilde{X})$.

Note that, if a single distribution $q$ is specified for variable $Y_i$, the only constraint imposed on the conditional distribution for $Y_i$ is:

$$p(Y_i = Y_{ij}|[\text{pa}(Y_i)]_k) = q(Y_i = Y_{ij}|[\text{pa}(Y_i)]_k).$$

## 5.2 LINEAR FRACTIONAL PROGRAMMING IN NATURAL EXTENSIONS

The objective here is to calculate posterior upper bounds (lower bounds are obtained by minimization):

$$\overline{p}(X_q|E) = \max \frac{\sum_{X_i \in (\tilde{X} \setminus \{X_q, E\})} p(\tilde{X})}{\sum_{X_i \in (\tilde{X} \setminus E)} p(\tilde{X})} \quad (3)$$

To guarantee that all credal sets contain valid distributions, the following unitary constraint must be added: $\sum_{\tilde{X}} p(\tilde{X}) = 1$.

The simplest natural extension is produced when no irrelevance relations are associated to a Quasi-Bayesian network [9]. In this case, the maximization in Expression (3), subject to linear constraints in Expressions (2) and the unitary constraint, is a *linear fractional program*. To guarantee that this linear fractional program has a solution, it is necessary to check that $\underline{p}(E)$ is non-zero; if $\underline{p}(E) = 0$, then the posterior lower envelope $\underline{p}(X_q|E)$ is also zero [28]. Linear fractional programs can be reduced to linear programs by a variety of methods [17, 23]; consequently, Quasi-Bayesian inferences (without irrelevance relations) can be solved by linear programming techniques.

## 5.3 REPRESENTATION OF IRRELEVANCE RELATIONS

Suppose that a variable $X_i$ is associated with a credal set $K(X_i|\text{pa}(X_i))$ and that the variables $W_i$ are judged irrelevant to $X_i$ given $\text{pa}(X_i)$. To represent the irrelevance relation, it is necessary to expand each constraint $C_l[p(X_i|\text{pa}(X_i))]$ into a family of constraints $C_l[p(X_i|\text{pa}(X_i), W_i = W_{ij})]$. Note that a new constraint is added for each value of $W_i$.

## 5.4 IRRELEVANCE CONSTRAINTS FOR NONDESCENDANTS

Consider the constraint that, for every variable $X_i$, nondescendants of $X_i$ are irrelevant to $X_i$ given the parents of $X_i$. For a variable $X_i$, denote the nondescendants of $X_i$ by $\text{nd}(X_i)$. Irrelevance constraints are satisfied by extending the replicating $C_l[p(X_i|[\text{pa}(X_i))]_k]$ for all the values of nondescendants $\text{nd}(X_i)$ such that $\text{pa}(X_i) = [\text{pa}(X_i)]_k$. Denote the set of constraints obtained in this manner by $C_l[p(X_i|\text{nd}(X_i))]$. By construction, if a joint distribution satisfies constraints $C_l[p(X_i|\text{nd}(X_i))]$, then it satisfies constraints $C_l[p(X_i|[\text{pa}(X_i)]_k)]$ (Appendix A.2).

Lower bounds are calculated by forming a linear fractional program with Expression (3) subject to linear constraints $C_l[p(X_i|\text{nd}(X_i))]$ and the unitary constraint. Even though irrelevance relations introduce a large number of constraints into this program, they also introduce simplifications into the problem, as demonstrated in the remainder of this section.

Consider a Quasi-Bayesian network where a group of variables $\tilde{Z}$ is associated with credal sets. Construct the set $\tilde{S}$ containing all variables in $\tilde{Z}$ and all variables that are predecessors of variables in $\tilde{Z}$. Call $\tilde{W}$ the set of all variables that are not in $\tilde{S}$.

**Theorem 2** *The calculation of Expression (3) can be*

*done by the solution of the program:*

$$\overline{p}(X_q|E) = \max \left( \frac{\sum_{Z_i \in (\tilde{S} \setminus \{X_q, E\})} q'(\tilde{S}) p(\tilde{S})}{\sum_{Z_i \in (\tilde{S} \setminus E)} q'(\tilde{S}) p(\tilde{S})} \right), \quad (4)$$

*subject to* $\sum_{\tilde{S}} p(\tilde{S}) = 1$ *and*

$$\sum_{j=1}^{|\hat{Z}_i|} \left( \gamma_{ijkl} \sum_{Z_m \in (\{\tilde{S}\} \setminus \{Z_i, nd(Z_i)\})} p(\tilde{S}) \right) - \quad (5)$$

$$\gamma_{i0kl} \sum_{Z_m \in (\tilde{S} \setminus nd(Z_i))} p(\tilde{S}) \leq 0,$$

*where the function* $q'$ *is:*

$$q'(\tilde{S}) = \sum_{W_i \in (\tilde{W} \setminus \{X_q, E\})} \left( \prod_{W_i \in \tilde{W}} q(W_i|pa(W_i)) \right).$$

*(Proof in Appendix A.3).*

The linear fractional program in this theorem is *not* a problem on variables $\tilde{X}$, but a reduced maximization problem where only the values for $p(\tilde{S})$ are free to vary. A standard Bayesian network algorithm generates $q'$ by essentially eliminating all variables in $\tilde{W}$.

The consequence of the theorem is that networks where most local credal sets are on the "top" of the graph can profit from irrelevance constraints. This is particularly promising in practical applications, because in general the most imprecise distributions are the priors, which are associated with nodes without parents.

### 5.5 REPRESENTATION OF INDEPENDENCE RELATIONS

To guarantee that $Y_i$ is independent of $X_i$ given $pa(X_i)$, it is necessary to enforce that: (1) $Y_i$ is irrelevant to $X_i$ given $pa(X_i)$, and (2) $X_i$ is irrelevant to $Y_i$ given $pa(X_i)$. The first constraint has been addressed in the previous paragraph, but the second constraint introduces new complexities into the problem. For example, suppose that a variable $X_5$ has variables $X_1, X_3, X_4$ as nondescendants, and $X_2$ as parent. The second irrelevance condition requires that the credal sets $K(X_1, X_3, X_4|X_2, X_5)$ and $K(X_1, X_2, X_4|X_2)$ contain the same functions. The difficulty is that neither of these credal sets is directly specified on the network; there is no simple constraint that ties them together.

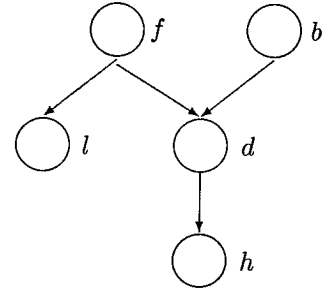

$$0.4 \leq p(f) \leq 0.5 \quad 0.4 \leq p(b) \leq 0.5$$
$$p(l|f) = 0.6 \quad p(d|f, b) = 0.8$$
$$p(l|f^c) = 0.05 \quad p(d|f, b^c) = 0.1$$
$$p(h|d) = 0.6 \quad p(d|f^c, b) = 0.1$$
$$p(h|d^c) = 0.3 \quad p(d|f^c, b^c) = 0.7$$

Figure 1: Example network (graphical structure and probabilistic statements).

### 5.6 INDEPENDENCE CONSTRAINTS FOR NONDESCENDANTS

Consider the constraint that, for every variable $X_i$, nondescendants of $X_i$ are independent from $X_i$ given the parents of $X_i$: The nondescendants of $X_i$ must be irrelevant to $X_i$, and $X_i$ must be irrelevant to its nondescendants given the parents of $X_i$. No efficient algorithm for inferences with such constraints is known; construction of a complex, non-linear optimization program is the only method that can be generally adopted at this point.

## 6 EXAMPLE

To illustrate the results and algorithms described previously, a simple example is discussed in this section. This example is based on the example described by Charniak [5] and on the calculations presented by Walley [28, Section 9.3.4].

Consider the graph in Figure 1. There are five binary variables in the graph (the superscript $c$ indicates negation). These relationships are summarized by the probabilistic model presented in Figure 1. Note that the probabilities for $f$ and $b$ are not specified exactly; instead, they are given as an interval $[0.4, 0.5]$. The question is how to evaluate the impact of this imprecision in probability values. To illustrate the various algorithms discussed in the paper, consider the calculation of $\underline{p}(d|l)$ and $\overline{p}(d|l)$.

**Type-1 extension** The simplest method to obtain the bounds is to identify the vertices of the local credal sets and generate a type-1 extension. The type-1 ex-tension has four vertices, because both the credal sets $K(f)$ and $K(b)$ have vertices $(0.4, 0.6)$ and $(0.5, 0.5)$. By calculating $p(d|l)$ for these four vertices, the bounds on $p(d|l)$ are obtained: the lower bound on $p(d|l)$ is 0.38615 and the upper bound is 0.44615.

**Natural extension without irrelevance relations**
If no irrelevance relation is stated concerning the network, then the expressions in Figure 1 and the unitary constraint are the only restrictions on the natural extension. To generate lower and upper bounds on $p(d|l)$, it is necessary to write these thirteen linear constraints (nine are equality constraints and four are inequality constraints) and solve a linear fractional program with the objective function $p(d,l)/p(l)$. The solution of this program produces the lower bound 0 and the upper bound 1 for $p(d|l)$, demonstrating that the absence of irrelevance relations can lead to inferences that are essentially vacuous.

**Natural extension with irrelevance relations**
Consider the effect of adding irrelevance relations, in particular the statement that the nondescendants of a variable are irrelevant to the variable given the parents of the variable. Four constraints represent this statement regarding credal sets: $0.4 \leq p(f|b) \leq 0.5$ and $0.4 \leq p(b|f) \leq 0.5$. To simplify the calculation of lower and upper bounds, Theorem 2 can be used. The upper bound is obtained by solving the program:

$$\max \frac{0.48 w_1 + 0.06 w_2 + 0.005 w_3 + 0.035 w_4}{0.6 w_1 + 0.6 w_2 + 0.05 w_3 + 0.05 w_4}$$

$$\text{s.t.} \begin{bmatrix} -3 & 0 & 2 & 0 \\ 2 & 0 & -2 & 0 \\ 0 & -3 & 0 & 2 \\ 0 & 2 & 0 & -2 \\ -3 & 2 & 0 & 0 \\ 2 & -2 & 0 & 0 \\ 0 & 0 & -3 & 2 \\ 0 & 0 & 2 & -2 \end{bmatrix} \times \begin{bmatrix} w_1 \\ w_2 \\ w_3 \\ w_4 \end{bmatrix} \leq \begin{bmatrix} 0 \\ 0 \\ 0 \\ 0 \\ 0 \\ 0 \\ 0 \\ 0 \end{bmatrix},$$

and $\sum w_i = 1$ ($w_1 = p(f,b)$, $w_2 = p(f,b^c)$, $w_3 = p(f^c,b)$, $w_4 = p(f^c,b^c)$). This program produces the upper bound 0.4509. By minimization, the lower bound 0.3818 is obtained. Note that these bounds are different from the bounds obtained by type-1 extension.

**Natural extension with independence constraints** The strongest statement considered here is the independence of a variable and its non-descendants given its parents. The natural extension is then derived from the full joint credal set $K(f,b)$, which has six vertices: $1/4(1,1,1,1)$, $(0.36, 0.24, 0.24, 0.16)$, $1/10(2,2,3,3)$, $1/10(2,3,2,3)$, $1/9(2,2,2,3)$, $1/11(2,3,3,3)$. Computation of $p(d|l)$ in each of the six joint distribution leads to the lower bound 0.3818 and the upper bound 0.4509.

## 7  CONCLUSION

The central contribution of this paper is the application of Walley's definitions of irrelevance and independence to the study of locally defined Quasi-Bayesian networks. The main technical contributions are novel algorithms for inference with natural extensions; research must now be conducted to limit the combinatorial explosion that occurs in the formulation of linear fractional programs for inferences with natural extensions. The paper also ties type-1 extensions to d-separation; this result provides a formal basis for the conceptual and computational attractiveness of type-1 extensions.

This paper focused on the calculation of upper bounds for the posterior probability of the event $\{X_q = X_{qj}\}$. Other problems can be solved using the same algorithms. For example, calculation of inferences for non-atomic events $A$ is immediate only by enlarging the summations that must be computed in the inference procedures. Algorithms presented in this paper also apply to calculation of lower and upper expectation, by enlarging summations and objective functions in linear programs.

The results presented in this paper pose an intellectually challenging question: Should we consider irrelevance or independence as a basic notion in the treatment of uncertainty? Both notions agree in standard probability theory, but they disagree in Quasi-Bayesian theory. Irrelevance is a more basic notion, as it can be used to define independence, and irrelevance judgements are less forceful than independence ones but still quite powerful. Should irrelevance be a more fundamental notion? This question can only be answered as research and applications are developed using Quasi-Bayesian models.

## A  PROOFS

### A.1  THEOREM 1

The following is a sketch for the proof of Theorem 1; a more detailed proof is available [8].

Consider three arbitrary disjoint sets of variables in the network, $\tilde{X}$, $\tilde{Y}$ and $\tilde{Z}$, such that $\tilde{X}$ is d-separated from $\tilde{Z}$ given $\tilde{Y}$. Take the type-1 extension $K(\tilde{X}, \tilde{Y}, \tilde{Z})$ and obtain, by conditionalization, $K(\tilde{X}|\tilde{Y}, \tilde{Z})$ and $K(\tilde{X}|\tilde{Y})$. Call ext$K$ the set of extreme points of $K$.

Given any function $f(\tilde{X})$ solely of $\tilde{X}$, obtain its lower expectation $\underline{E}[f(\tilde{X})|\tilde{Y}, \tilde{Z}] =$

$\min_{p \in \text{ext} K(\tilde{X}|\tilde{Y},\tilde{Z})} \left( \sum_{\tilde{X}} f(\tilde{X}) p(\tilde{X}|\tilde{Y},\tilde{Z}) \right)$. The minimum is attained at an extreme point of the type-1 extension. Because every such extreme point satisfies Expression (1), $p(\tilde{X}|\tilde{Y},\tilde{Z}) = p(\tilde{X}|\tilde{Y})$ for these points (by d-separation), and the lower expectation is equal to $\underline{E}[f(\tilde{X})|\tilde{Y}]$.

Because a lower expectation uniquely defines a convex set of distributions (Section 2.2), the lower expectation $\underline{E}[f(\tilde{X})|\tilde{Y}]$ uniquely defines $K(\tilde{X}|\tilde{Y})$ and the lower expectation $\underline{E}[f(\tilde{X})|\tilde{Y},\tilde{Z}]$ uniquely defines $K(\tilde{X}|\tilde{Y},\tilde{Z})$. Because both lower expectations are equal for arbitrary $f$, the underlying credal sets are the same. This argument guarantees that $\tilde{Z}$ is irrelevant to $\tilde{X}$ given $\tilde{Y}$; the same argument proves that $\tilde{X}$ is irrelevant to $\tilde{Z}$ given $\tilde{Y}$. So $\tilde{X}$ is independent of $\tilde{Z}$ given $\tilde{Y}$.

## A.2 RELEVANT LEMMAS

The following result is used in Section 5.4:

**Lemma 1** *If a joint distribution satisfies constraints $C_l[p(X_i|nd(X_i))]$, then it satisfies constraints $C_l[p(X_i|pa(X_i))]$.*

To prove this result, take $\tilde{W}(X_i) = \emptyset$ in the following theorem.

**Lemma 2** *Consider a joint distribution that satisfies constraints $C_l[p(X_i|nd(X_i))]$, and for every node $X_i$, $\tilde{W}(X_i)$ is a subset of $nd(X_i)$ that does not overlap with the parents of $X_i$. Then the following constraints also satisfied:*

$$\sum_{j=1}^{|\hat{X}_i|} \gamma_{ijkl} p(X_i = X_{ij}|[pa(X_i)]_k, \tilde{W}(X_i)) \leq \gamma_{i0kl}. \quad (6)$$

*Sketch of proof.* Consider an arbitrary joint distribution satisfying constraints $C_l[p(X_i|nd(X_i))]$. Denote the set $(nd(X_i)\backslash\{pa(X_i), \tilde{W}(X_i)\})$ by $\tilde{W}'(X_i)$. Obtain by marginalization the distribution of $\tilde{W}'(X_i)$, $p(\tilde{W}'(X_i))$.

Select all constraints that are repetitions of a single original constraint for fixed $[pa(X_i)]_k$. These constraints are all identical, except that values of $\tilde{W}(X_i)$ and $\tilde{W}'(X_i)$ vary across constraints. Multiply every one of these constraints by the appropriate value of $p(\tilde{W}'(X_i))$, and add all constraints that refer to a particular value of $\tilde{W}(X_i)$; constraints (6) are then obtained after algebraic manipulations.

## A.3 THEOREM 2

First note that the linear fractional program in the statement of the theorem is identical to the following program:

$$\overline{p}(X_q|E) = \max \left( \frac{\sum_{X_i \in (\tilde{X}\backslash\{X_q,E\})} p(\tilde{X})}{\sum_{X_i \in (\tilde{X}\backslash E)} p(\tilde{X})} \right), \quad (7)$$

subject to constraints (5), $\sum_{\tilde{S}} p(\tilde{S}) = 1$ and $p(\tilde{X}) = q(\tilde{W}|\tilde{S})p(\tilde{S})$, where

$$q(\tilde{W}|\tilde{S}) = \prod_{W_i \in \tilde{W}} q(W_i|pa(W_i)). \quad (8)$$

Note that $q(\tilde{W}|\tilde{S})$ is the unique joint distribution for $\tilde{W}$ given $\tilde{S}$. Uniqueness is guaranteed by the fact that the variables in $\tilde{W}$ form a Bayesian network: (1) irrelevance is equal to independence in standard Bayesian networks; (2) Lemma 2 guarantees that all irrelevance conditions are valid when restricted to the network of $\tilde{W}$; (3) independence of a variable from its nondescendants given its parents characterizes a unique Bayesian network [22].

The strategy of the proof is to demonstrate that the linear program expressed by (7) subject to constraints (5), (8) and $\sum_{\tilde{S}} p(\tilde{S}) = 1$, is identical to program (3) subject to $C_l[p(X_i|nd(X_i))]$ and the unitary constraint.

Start from program (3). Uniqueness of $q(\tilde{W}|\tilde{S})$ leads to constraints:

$$p(\tilde{W}|\tilde{S}) = q(\tilde{W}|\tilde{S}),$$

which are equivalent to the constraints summarized by Expression (8). Use this equality in Expressions $C_l[p(X_i|nd(X_i))]$ and the unitary constraint. Expression $\sum_{\tilde{S}} p(\tilde{S}) = 1$ is immediately obtained from the unitary constraint. For constraints $C_l[p(X_i|nd(X_i))]$, divide $\tilde{W}$ in two sets of variables; $\tilde{W}'$ contains variables in $\tilde{W}$ that are nondescendants of $X_i$, and $\tilde{W}''$ contains variables in $\tilde{W}$ that are descendants of $X_i$. Constraints $C_l[p(X_i|nd(X_i))]$ become:

$$\sum_{j=1}^{|\hat{X}_i|} \left( \gamma_{ijkl} \sum_{\bar{X}\backslash\{X_i, nd(X_i)\}} q(\tilde{W}''|\tilde{S},\tilde{W}')q(\tilde{W}'|\tilde{S})p(\tilde{S}) \right)$$
$$- \gamma_{i0kl} \sum_{\bar{X}\backslash nd(X_i)} q(\tilde{W}''|\tilde{S},\tilde{W}')q(\tilde{W}'|\tilde{S})p(\tilde{S}) \leq 0.$$

The summations involve all variables in $\tilde{W}''$, so these variables can be summed out. Variables in $\tilde{W}'$ are fixed and make no reference to $X_i$ or any of its descendants, so they can be taken out of the summation and cancelled. These operations reduce the inequality above to constraint (5). The only situation where this cancellation cannot occur is when a node has no nondescendants; in this case, all other nodes are descendants of the node and are summed out so the result holds. This proves that program (7) subject to (5), (8), and $\sum_{\tilde{S}} p(\tilde{S}) = 1$, is identical to program (3) subject to $C_l[p(X_i|\mathrm{nd}(X_i))]$ and the unitary constraint.

## Acknowledgements

I thank my former advisor, Eric Krotkov, for substantial support during the research that led to this work.

## References

[1] J. O. Berger. Robust Bayesian analysis: Sensitivity to the prior. *Journal of Statistical Planning and Inference*, 25:303–328, 1990.

[2] J. S. Breese and K. W. Fertig. Decision making with interval influence diagrams. *Uncertainty in Artificial Intelligence 6*, pages 467–478. Elsevier Science, North-Holland, 1991.

[3] A. Cano, J. E. Cano, and S. Moral. Convex sets of probabilities propagation by simulated annealing. *Fifth IPMU*, pages 4–8, July 1994.

[4] J. Cano, M. Delgado, and S. Moral. An axiomatic framework for propagating uncertainty in directed acyclic networks. *International Journal of Approximate Reasoning*, 8:253–280, 1993.

[5] E. Charniak. Bayesian networks without tears. *AI Magazine*, pages 50–63, Fall 1991.

[6] L. Chrisman. Independence with lower and upper probabilities. *XII Uncertainty in Artificial Intelligence Conference*, pages 169–177, 1996.

[7] L. Chrisman. Propagation of 2-monotone lower probabilities on an undirected graph. *XII Uncertainty in Artificial Intelligence Conference*, pages 178–186, 1996.

[8] F. Cozman. Independence relations in the robustness analysis of multivariate probabilistic models. Submitted to the XII Conferência Brasileira de Automática, Brasil, 1998 (available from the author).

[9] F. Cozman. Robustness analysis of Bayesian networks with local convex sets of distributions. *XIII Uncertainty in Artificial Intelligence Conference*, 1997.

[10] L. de Campos and S. Moral. Independence concepts for convex sets of probabilities. *XI Uncertainty in Artificial Intelligence*, 1995.

[11] T. L. Fine. Lower probability models for uncertainty and nondeterministic processes. *Journal of Statistical Planning and Inference*, 20:389–411, 1988.

[12] J. Gebhardt and R. Kruse. Learning possibilistic networks from data. *Fifth International Workshop on Artificial Intelligence and Statistics*, 1995.

[13] D. Geiger, T. Verma, and J. Pearl. d-separation: from theorems to algorithms. *Uncertainty in Artificial Intelligence 5*, 1990.

[14] F. J. Giron and S. Rios. Quasi-Bayesian behaviour: A more realistic approach to decision making? *Bayesian Statistics*, pages 17–38. University Press, Valencia, Spain, 1980.

[15] H. E. Kyburg Jr. Bayesian and non-Bayesian evidential updating. *Artificial Intelligence*, 31:271–293, 1987.

[16] J. Y. Halpern and R. Fagin. Two views of belief: Belief as generalized probability and belief as evidence. *Artificial Intelligence*, 54:275–317, 1992.

[17] T. Ibaraki. Solving mathematical programming problems with fractional objective functions. *Generalized Concavity in Optimization and Economics*, pages 440–472. Academic Press, 1981.

[18] F. V. Jensen. *An Introduction to Bayesian Networks*. Springer Verlag, New York, 1996.

[19] J. B. Kadane. *Robustness of Bayesian Analyses*, volume 4 of *Studies in Bayesian econometrics*. Elsevier Science Pub. Co., New York, 1984.

[20] I. Levi. *The Enterprise of Knowledge*. The MIT Press, Cambridge, Massachusetts, 1980.

[21] J. Pearl. On probability intervals. *International Journal of Approximate Reasoning*, 2:211–216, 1988.

[22] J. Pearl. *Probabilistic Reasoning in Intelligent Systems: Networks of Plausible Inference*. Morgan Kauffman, San Mateo, CA, 1988.

[23] S. I. Schaible and W. T. Ziemba. *Generalized Concavity in Optimization and Economics*. Academic Press, 1981.

[24] T. Seidenfeld, M. J. Schervish, and J. B. Kadane. A representation of partially ordered preferences. *The Annals of Statistics*, 23(6):2168–2217, 1995.

[25] P. P. Shenoy and G. Shafer. Axioms for probability and belief-function propagation. *Uncertainty in Artificial Intelligence 4*, pages 169–198. Elsevier Science Publishers, North-Holland, 1990.

[26] E. H. Shortliffe and B. G. Buchanan. *Rule-based expert systems*. The Addison-Wesley series in artificial intelligence. Addison-Wesley, Reading, Mass., 1985.

[27] B. Tessem. Interval probability propagation. *International Journal of Approximate Reasoning*, 7:95–120, 1992.

[28] P. Walley. *Statistical Reasoning with Imprecise Probabilities*. Chapman and Hall, New York, 1991.

[29] L. Wasserman. Recent methodological advances in robust Bayesian inference. *Bayesian Statistics 4*, pages 483–502. Oxford University Press, 1992.